\documentclass[conference]{IEEEtran}
\IEEEoverridecommandlockouts
\usepackage{cite}
\usepackage{amsmath,amssymb,amsfonts}
\usepackage{algorithmic}
\usepackage{graphicx}
\usepackage{textcomp}
\usepackage{xcolor}
\usepackage{tabu} 
\usepackage{amsbsy}
\usepackage{atbegshi}
\def\BibTeX{{\rm B\kern-.05em{\sc i\kern-.025em b}\kern-.08em
    T\kern-.1667em\lower.7ex\hbox{E}\kern-.125emX}}
\begin{document}

\title{An Automated Machine Learning Approach to Inkjet Printed Component Analysis: A Step Toward Smart Additive Manufacturing\\
}

\author{ Abhishek Sahu$^1$, Peter H. Aaen$^2$ and Praveen Damacharla$^3$\\
$^1$Qorvo Us Inc., Greensboro, NC, USA\\
$^2$Colorado School of Mines, Golden, CO, USA\\
$^3$KineticAI Inc., The Woodlands, TX, USA\\
$^1$abhishek.sahu@ieee.org, $^2$paaen@mines.edu,
$^3$praveen@kineticai.com,
}

\maketitle

\begin{abstract}
\looseness=-1 In this paper, we present a machine learning based architecture for microwave characterization of inkjet printed components on flexible substrates. Our proposed architecture uses several machine learning algorithms and automatically selects the best algorithm to extract the material parameters (ink conductivity and dielectric properties) from on-wafer measurements. Initially, the mutual dependence between material parameters of the inkjet printed coplanar waveguides (CPWs) and EM-simulated propagation constants is utilized to train the machine learning models. Next, these machine learning models along with measured propagation constants are used to extract the ink conductivity and dielectric properties of the test prototypes. To demonstrate the applicability of our proposed approach, we compare and contrast four heuristic based machine learning models. It is shown that eXtreme Gradient Boosted Trees Regressor (XGB) and Light Gradient Boosting (LGB) algorithms perform best for the characterization problem under study. 
\end{abstract}


\section{Introduction}
\looseness=-1 Printed electronics has gathered wide attention in the emerging markets such as internet of things as it offers biodegradable and cost-effective solutions. Recently, many traditional methods such as gravure printing, flexography, screen printing, and inkjet printing have been introduced in manufacturing of microwave circuits \cite{c1}. Among them, inkjet printing 
is most popular in the scientific and industrial community due to its attractive features including low manufacturing cost, large-area processability and lower carbon footprints. RF/microwave community has reported numerous applications of inkjet printing including antennas, wireless power transfer topologies, sensors, and microwave components \cite{c2,c3,c4,c5} for low-loss and high-speed
communication systems.

\begin{figure}[t]
	\centering
	\includegraphics[width=0.85\columnwidth]{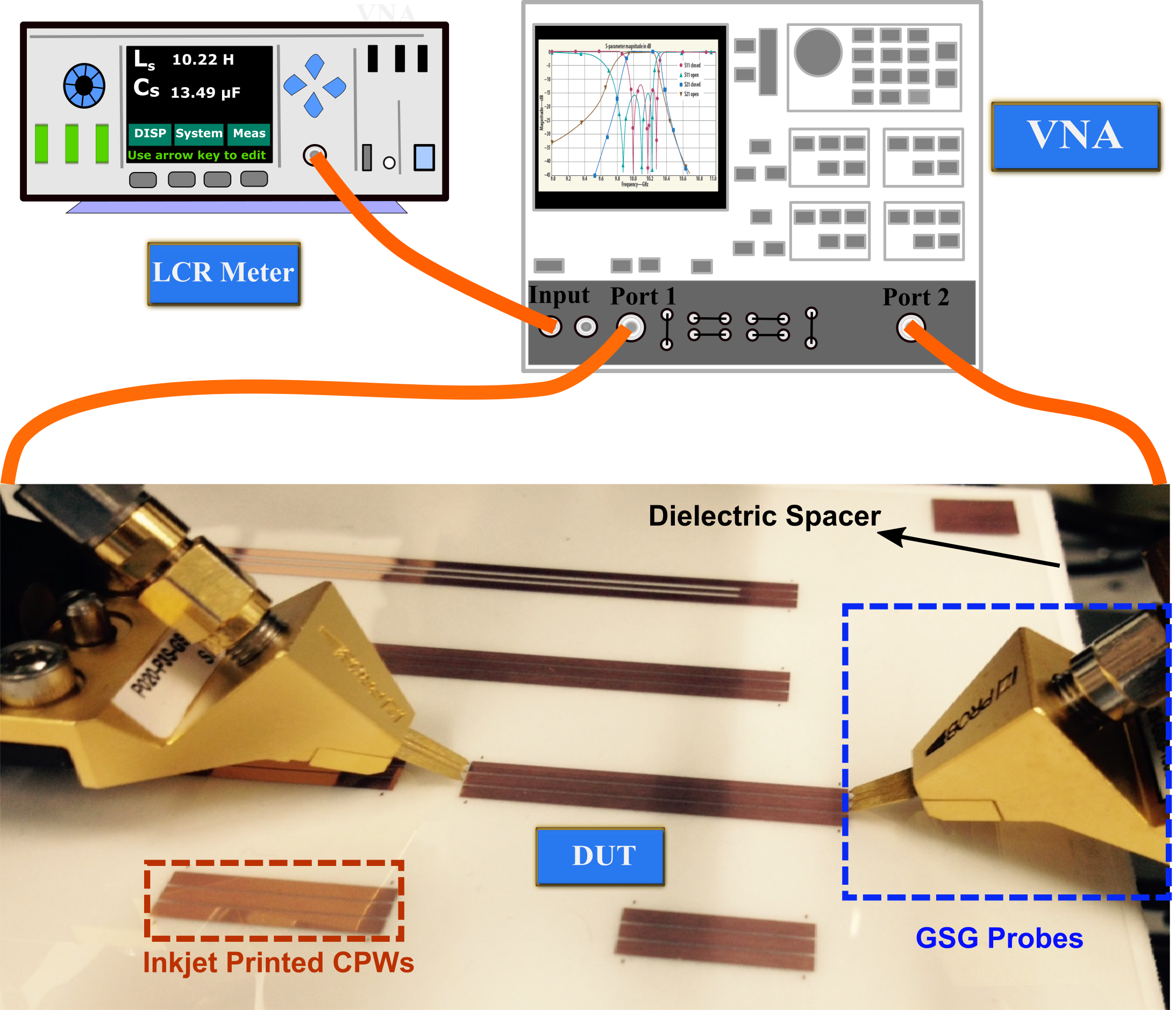}
	\caption{\small{Measurement setup for the inkjet printed CPWs. GSG probes are connected to measured the small signal S-parameters of the printed CPWs. An LCR meter is connected in series to measure the dc resistance of the THRU line. }}
	\label{f:Test_figure2}
\end{figure}

\looseness=-1 Significant effort has been devoted from the research community to extract the material parameters of inkjet printed components from their measurements \cite{c6}-\cite{c10}. Conventional approaches need high precision profile measurements to extract the ink conductivity and a conformal mapping method (CMM) based on elliptical integrals to evaluate the dielectric constants. However, as these analytical methods are manually calculated to extract the parameters independently, they require rigorous human effort. Recently, computer-aided design (CAD) based-techniques utilizing commercial EM simulation packages and measurements have been proposed for automation and concurrent parameter extraction \cite{c10,c11,c12,c13}. The problem with this approach is it requires a large number of iterative simulations. Further, as the number of variables increase, the optimization space becomes very complex and sometimes may lead to convergence problems. 
Therefore, there is a need for an alternative characterization approach that offers a more straightforward automation and is able to extract multiple parameters simultaneously without increasing process complexity.

\looseness=-1 Recently machine learning based approaches have gained wide attention as they make the characterization problem computationally intelligent \cite{c14,c15}. However, machine learning based approaches require significant expertise. Additionally, extensive human effort is needed to select an appropriate algorithm and their hyper parameters for each problem.  To overcome aforementioned problems, we present an automated machine learning models (AutoML) for accurate extraction of the material parameters of inkjet printed components. The proposed approach is fast and robust and is able to obtain all of the material parameters simultaneously from a single set of measurements. The proposed AutoML architecture tries numerous combination of machine learning algorithms for the training of each parameter and consecutively selects the best algorithm for each parameter. We demonstrate that the proposed architecture is able to predict the material properties with only a few sample of measured data.


\looseness=-1 The paper is organized as follows. Section II presents the detailed procedure to extract the propagation constant from measured S-parameters and a multiline TRL calibration. Section III presents the details of the machine learning models. Section IV presents the material characterization using machine learning models and compares the performance of the proposed models. We finally conclude the paper in section V.

\section{Data Extraction}
\subsection{Fabrication and Measurement}
\looseness=-1 To demonstrate the applicability of our proposed approach, we fabricated a set of CPW lines using a commercial process with Inkjetflex \cite{c16}. The details of fabrication can be found in \cite{c11}. 
The CPW lines (Fig. 1) have a 1.983 mm center conductor (\textit{$W_{c}$}), a 0.13 mm gap (\textit{g}), and 1.983 mm wide ground planes (\textit{$W_{g}$}) fabricated on 125-$\mu$m-thick ($t_{FS}$) PET substrate with approximately 2 $\mu$m copper plating. To complete a benchmark multiline TRL calibration, the fabrication included a set of 6 CPWs with lengths of \textit{l}=\big (14.97 mm, 18.42 mm, 23.57 mm, 35.78 mm, 61.60 mm, 97.93 mm\big), a symmetric short circuit and a reflect standard. The reflect standard consisted of a 6.5-mm-long transmission line on each port centered about the termination. 

\looseness=-1 Fig. 1 shows the measurement setup for the inkjet printed CPWs. An Agilent E8364B vector network analyzer and an on-wafer probing system were used to carry out S-parameter measurements from 10 MHz to 20 GHz. The probing system includes two ground-signal-ground microwave probes (1000-$\mu$m pitch), probe manipulators, and an optical microscope. 
The power level of VNA was set to -17.0 dBm. An HP 4285A LCR meter was connected with the VNA to measure the dc resistance of the THRU. During measurements, the CPWs were supported by a 0.5-cm-thick ($t_{DS}$) porous, dielectric spacer to mitigate any potential deflection of the flexible PET substrate and avoid propagation of any undesired parallel-plate and microstrip modes due to the metallic vacuum chuck. 

\begin{figure}[t]
	\centering
	\includegraphics[width=0.85\columnwidth]{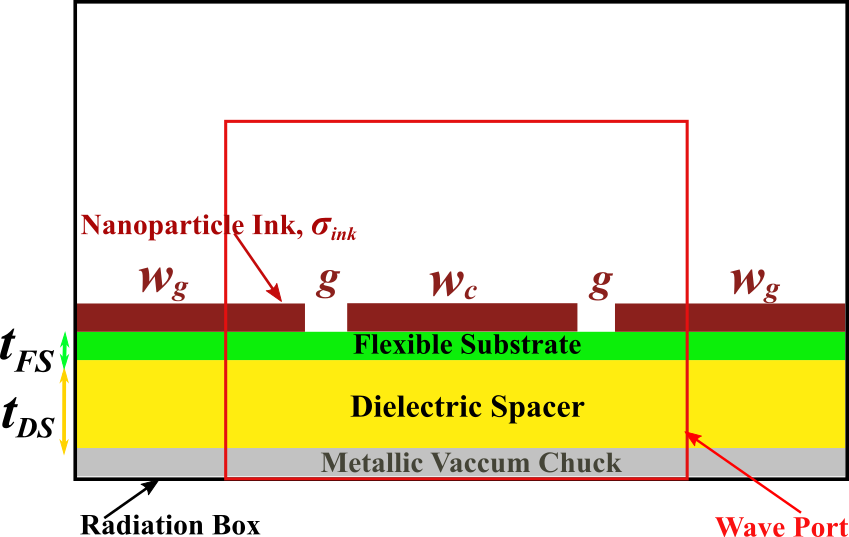}
	\caption{\small{Crosssection of CPWs in a CAD environment. Silver nanoparticle inks of conductivity ($\sigma_{ink}$) are deposited on a (\textit{$t_{FS}$}) thick flexible substrate to print CPWs with a ground plane of width (\textit{$W_{g}$}), center conductor of width (\textit{$W_{c}$}) and gap of width (\textit{g}). The CPWs are supported by a dielectric spacer of thickness (\textit{$t_{DS}$}) underneath.}} 
	\label{f:Test_figure2}
\end{figure}

\looseness=-1 We employed a multiline TRL calibration to extract the complex propagation constant from $uncalibrated$ S-parameter measurements. It should be noted that, in multiline TRL, the calibration is defined relative
to the characteristic impedance of the transmission line ($Z_0$). Additionally, the propagation constant is extracted by analytically solving the eigenvalues of the measured cascade matrices as \cite{c17}
\begin{equation}
\lambda_1^{ij}, \lambda_2^{ij}= \dfrac{1}{2}\bigg[\big(M_{11}^{ij}+M_{22}^{ij}\big) \pm \sqrt{\big(M_{11}^{ij}-M_{22}^{ij}\big)^2 + 4M_{12}^{ij}M_{21}^{ij}}\bigg]
\end{equation}
where $M_{12}^{ij}$ represent the measured cascade matrices of line pair $i$ and $j$. For a two port device, the relationship between S and M parameters can be obtained as
\begin{equation}
   M= \dfrac{1}{S_{21}}
   \begin{bmatrix}
     \big(S_{12}S_{21}-S_{11}S_{22}\big) & S_{11}\\
    -S_{22} & 1\\
     \end{bmatrix}.
\end{equation}
The accuracy of the multiline TRL calibration can be validated by a calibration comparison method as described in \cite{c17}. 

\begin{figure*}
	\centering
	\includegraphics[width=0.85\textwidth]{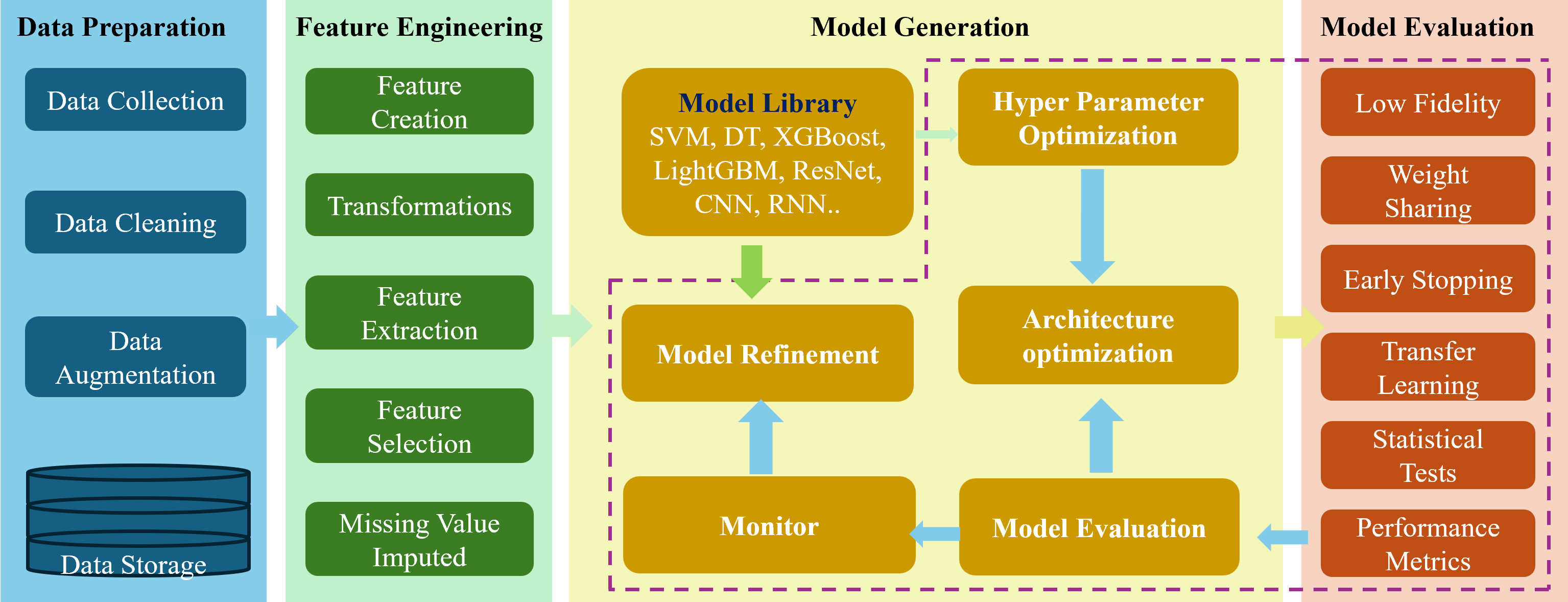}
	\caption{\small{Architecture of the proposed automatic machine learning (AutoML). }}
	\label{f:Test_figure3}
\end{figure*}

\begin{figure*}
	\centering
	\includegraphics[width=0.82\textwidth]{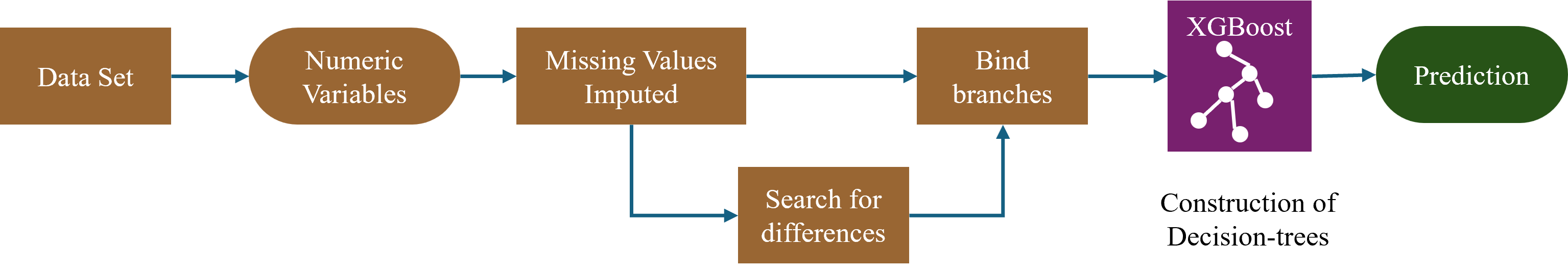}
	\caption{\small{eXtreme gradient boosted trees regressor with early stopping (learning rate = 0.02. }}
	\label{f:Test_figure4}
\end{figure*}

\begin{figure*}
	\centering
	\includegraphics[width=0.82\textwidth]{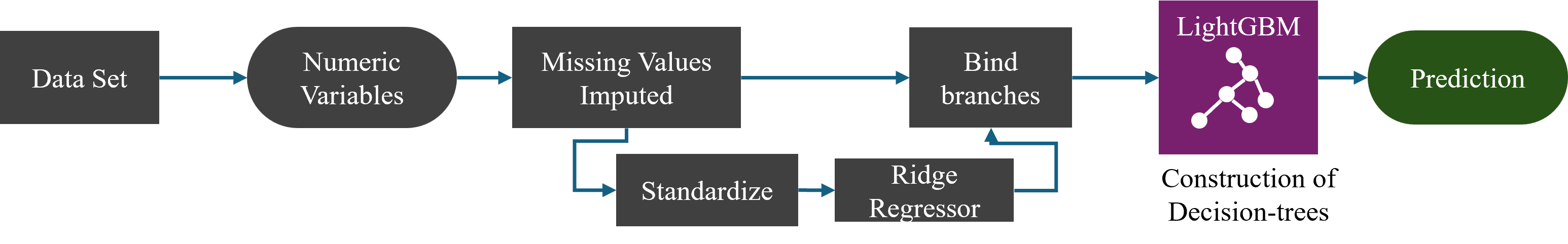}
	\caption{\small{Light gradient boosting on elasticnet predictions.}}
	\label{f:Test_figure5}
\end{figure*}

\subsection{EM Modeling}
\looseness=-1 In order to obtain the training data for the machine learning models, full-wave finite element EM simulations were performed for the inkjet printed CPWs. Fig. 2 shows the cross section of the CPWs in an EM simulation environment. The full-wave EM simulation accounts for all terms of Maxwell's equations and thus takes into account radiation losses in $\alpha$. Single mode propagation is crucial for multiline TRL calibration. Therefore, to emulate the measurement setup, we ensured that the propagating EM field in the finite element model contains only one mode and no higher order modes are present in the waveguide ports. Once the EM simulation set up is validated, parametric analysis was performed by sweeping each of the input parameters ($\sigma_{ink}$, $\epsilon_{FS}$, $\epsilon_{DS}$, $tan \delta$) to obtain the corresponding output parameters ($\alpha$ and $\beta$). These data is then utilized to train the machine learning models.

\section{AutoML Methodology}
\looseness=-1 Fig. \ref{f:Test_figure3} presents the proposed  AutoML architecture for smart characterization of inkjet printed components. 
The architecture leverages a suite of regression models and unfolds in several systematic stages, starting from data preprocessing to model training and evaluation, ensuring a comprehensive analysis of the inkjet printed components. 

\subsection{Data Processing}
The data preparation is the first step of our AutoML architecture and comprises of several steps as follows:
\begin{itemize}
{
\item Data Collection: The training data includes frequency dependent $\alpha$ and $\beta$ by varying the material parameters ($\sigma_{ink}$, $\epsilon_{DS}$, $\epsilon_{FS}$ and $tan \delta$). 

\item Data Description: The dataset consists of a substantial 47,200 data points for each parameter, with the frequency spanning a broad spectrum from 10 MHz to 20 GHz. 

\item Data Cleaning: We undertook a rigorous cleaning process to ensure elimination of discrepancies, duplicate values, thereby enhancing the overall quality of our data.

\item Data Augmentation: We employed data augmentation techniques to synthesize new derivative data points, which in turn provides a more diverse and comprehensive dataset for training our models.


\item Data Selection and Partitioning: We employed two distinct methods for partitioning the dataset. The first method involved a random selection process where we allocated 75\% of the data for training, 5\% for testing, and the remaining 20\% for validation. The second method adjusted the distribution to utilize 90\% of the data for training, with the remaining 10\% split evenly between testing and validation, each receiving 5\%. 
}
\end{itemize}

\subsection{Feature Engineering}
This stage is crucial for enhancing the model's predictive power and involves:
\begin{itemize}

\item Feature Creation: Generating new features from the existing dataset 
\item Transformations: Applying mathematical or statistical transformations to normalize or scale the data features.
\item Feature Extraction: Deriving new features from the raw data that are more informative and non-redundant.
\item Feature Selection: Selecting crucial features to reduce dimensionality and improve model performance.
\item Missing Value Imputation: Handling missing data by employing techniques to input or generate substitute values.

\end{itemize}

\subsection{Model Training}
\looseness=-1 Subsequent to data preprocessing, the architecture embarks on the core phase of model training and evaluation. This phase employs a diverse array of 39 regression models, ensuring a comprehensive exploration of the solution space. The models include, but are not limited to, Decision Tree (DT), XgBoost (Fig. \ref{f:Test_figure4}), LightGBM (Fig. \ref{f:Test_figure5}), and ResNet. The core of the AutoML architecture revolves around generating and refining machine learning models:

\begin{itemize}
\item Model Library: A repository of various machine learning algorithms, such as SVM, Decision Trees, XGBoost, LightGBM, ResNet, CNN, RNN, etc., 
\item Hyperparameter Optimization: An iterative process of tuning the hyperparameters to find the most effective model configurations.
\item Architecture Optimization: Fine-tuning the structure or architecture of models for optimal performance.
\item Model Refinement: Iterative process for improving models based on the evaluation metrics and feedback.
\item Monitoring: Continuous observation of model performance during the training and validation process.
\end{itemize}

\looseness=-1 The architecture emphasizes a feedback loop where the model evaluation enables further model generation, refinement, and feature engineering, ensuring a dynamic and adaptive AutoML system. 
In the Fig. \ref{f:Test_figure4} and \ref{f:Test_figure5} we see an exemplified workflow for a dataset undergoing processing, such as imputation of missing values, standardization, and subsequent training of machine learning models like XGBoost and LightGBM leading to predictions. This indicates a parallel processing and ensemble approach, possibly combining predictions from multiple models to enhance the final prediction performance. The presented AutoML architecture illustrates a sophisticated, iterative, and integrative approach to developing machine learning models. 

\subsection{Comparative Analysis and Model Selection}
Upon the completion of training and evaluation, the architecture conducts a comparative analysis of the models. This analysis is pivotal for choosing the model that exhibits the most promising results in the context of smart additive manufacturing and includes: 
\begin{itemize}
\item Low Fidelity: Rapid, approximate evaluations of model performance through training on subsets of data or reduced epochs.
\item Weight Sharing: Utilizing shared weights across different models to reduce computational overhead.
\item Early Stopping: Preventing overfitting by halting the training process when the model's performance on the validation set plateaus or decreases.
\item Transfer Learning: Applying knowledge from one domain to improve performance in another, often speeding up the training process.
\item Statistical Tests: Using hypothesis testing to ascertain the statistical significance of the observed differences in model performance.
\item Performance Metrics: Evaluating models using various metrics such as accuracy, recall, AUC-ROC for classification, and MSE, RMSE for regression
\end{itemize}

\looseness=-1 The scatter plots shown in Fig. 6 delineate the predictive performance of various machine learning models, specifically DT, LightGBM, XGBoost, and ResNet, across four distinct parameters under study. The plots illustrate a comparative analysis between the actual and predicted values, where a perfect prediction would align with the blue line. For the parameter $\epsilon_{DS}$, the XGBoost and LightGBM models demonstrate a tight clustering around the line of perfect prediction, indicating a robust predictive capability, while the DT model exhibits greater dispersion, signifying potential over-fitting or model variance. This trend is echoed in the $\epsilon_{FS}$ parameter, where all models show commendable predictive accuracy, although the DT model seems slightly more scattered compared to the ensemble methods. Interestingly, in the assessment of $tan \delta$, the LightGBM and XGBoost models maintain their high fidelity to the actual values, with the DT model's predictions diverging more noticeably at lower actual values, potentially indicating a lack of sensitivity to finer variations within the dataset.

\looseness=-1 The inclusion of the ResNet model in the $\sigma_{ink}$ parameter plot reveals its competitive performance alongside XGBoost, both adhering closely to the line of unity across a broad range of values, albeit with some deviation at higher magnitudes. The DT model, while generally following the trend, shows substantial variance at the higher end of the scale, suggesting that while it captures the overall pattern, its precision may be compromised under conditions of increased parameter values. 


\begin{figure}
	\centering
	\includegraphics[width=0.95\columnwidth]{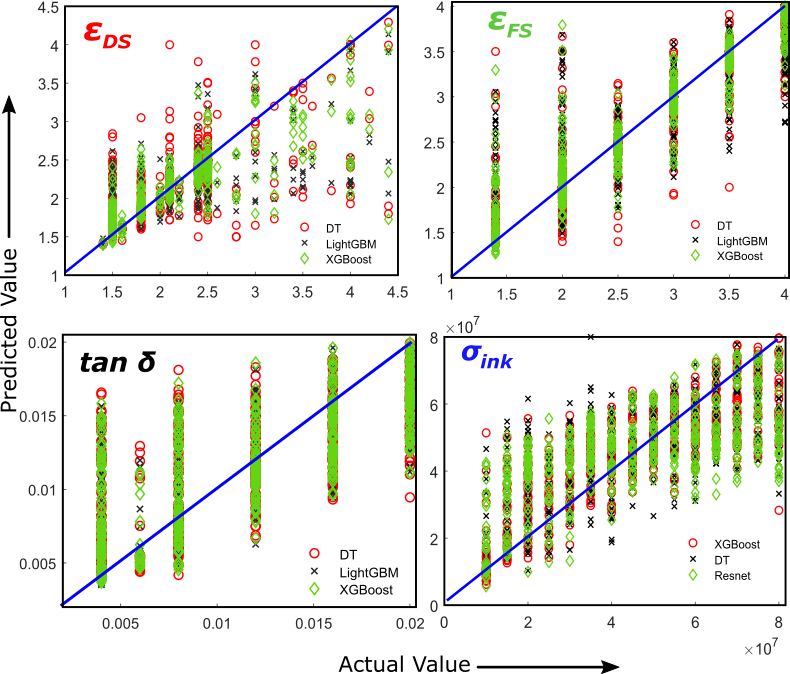}
	\caption{\small{Validation of training with holdout samples for retrieved material parameters using different models.}} 
	\label{f:Test_figure6}
\end{figure}

\begin{table}[t]
    \renewcommand{\arraystretch}{1.3}
	\centering
    \centering 
    \caption{Model performance metrics: RMSE Validation Score}
    \begin{tabular}{|l|r|r|r|r|}
    \hline
    \textbf{Model Name} & $\pmb{\sigma_{ink}}$ & $\pmb{\epsilon_{DS}}$ & $\pmb{\epsilon_{FS}}$ & $\pmb{tan \delta}$ \\ \hline
    XGBoost & 0.2573 & 0.306 & 0.2573 & 0.00379 \\ \hline
    LightGBM & 0.00378 & 0.4171 & 0.4171 & 0.00379 \\ \hline
    DT & 0.3492 & 0.4371 & 0.3492 & 0.00395 \\ \hline
    ResNet & 7523400 & 0.4925 & 0.4358 & 0.00406 \\ \hline
    \end{tabular}
    \label{table:models1}
\end{table}

\looseness=-1 To further elaborate the comparative analysis, we report the root mean square error(RMSE) validation and holdout scores for regression models in Table \ref{table:models1} and Table \ref{table:models2}. The XGBoost model showcases commendable consistency, as indicated by identical $\sigma_{ink}$ and $\epsilon_{FS}$ scores of 0.2573, coupled with a relatively low RMSE validation score $tan \delta$ of 0.00379, suggesting a strong predictive accuracy with minimal overfitting. The LightGBM algorithm, while demonstrating a minuscule $\sigma_{ink}$ coefficient of 0.00378, unexpectedly reports high $\epsilon_{DS}$ and $\epsilon_{FS}$ values at 0.4171, which, despite sharing an identical RMSE validation score with XGBoost, might point to an overestimation in prediction under certain data distributions or a potential overfitting that the RMSE score alone does not fully reveal.

\looseness=-1 The DT model presents a higher $\sigma_{ink}$ coefficient and $\epsilon_{FS}$ value at 0.3492 with the highest $\epsilon_{DS}$ coefficient at 0.4371, aligning with a slightly increased RMSE validation score of 0.00395. This increment in RMSE suggests that the DT model captures a broader range of data patterns at the expense of prediction precision. The ResNet model, primarily recognized for its applications in deep learning, displays an excessively high $\sigma_{ink}$ coefficient of 7523400, which is indicative of overfitting, especially when considered alongside the highest RMSE Validation Score of 0.00406 in the group. Its $\epsilon_{DS}$ and $\epsilon_{FS}$ scores of 0.4925 and 0.4358, respectively, though on the higher side, are overshadowed by the disproportionate $\sigma_{ink}$ value, suggesting that the model's complexity may not be justified for the given regression task \cite{c18}.

\begin{table}[t]
    \renewcommand{\arraystretch}{1.3}
	\centering
    \centering 
    \caption{Model performance metrics: RMSE Holdout Score}
    \begin{tabular}{|l|r|r|r|r|}
    \hline
    \textbf{Model Name} & $\pmb{\sigma_{ink}}$ & $\pmb{\epsilon_{DS}}$ & $\pmb{\epsilon_{FS}}$ & $\pmb{tan \delta}$ \\ \hline
    XGBoost & 0.2582 & 0.3146 & 0.2582 & 0.00375 \\ \hline
    LightGBM & 0.00376 & 0.4162 & 0.4162 & 0.00377 \\ \hline
    DT & 0.3459 & 0.4517 & 0.3459 & 0.00389 \\ \hline
    ResNet & 17025000 & 0.4898 & 0.4361 & 0.00401 \\ \hline
    \end{tabular}
    \label{table:models2}
\end{table}

\looseness=-1 In assessing the efficacy of various machine learning models for regression tasks shown in Table \ref{table:models2}, we observe a diverse range of outcomes as illustrated in the table of results. The XGBoost model exhibits a moderate level of precision 
indicating a relatively consistent and reliable prediction capability. In contrast, the RSME holdout scores for LightGBM model
suggest that it may be more sensitive to the specific dataset or require careful tuning of parameters to achieve optimal performance. The DT model shows a higher variation in coefficients. 
This could imply a model is more prone to overfitting or that it captures more complexity in the data. Lastly, the ResNet architecture, typically used for deep learning tasks, presents an anomaly in its $\sigma_{ink}$ coefficient at 17025000, a clear indication of overfitting.
This substantial $\sigma_{ink}$ value might be a result of the model's complexity and a large number of parameters, which could have led to capturing noise in the training data. 

\looseness=-1 This comparative evaluation reveals that while XGBoost and LightGBM maintain a lower RMSE, indicating superior generalization, the DT shows a slightly reduced predictive precision, and the ResNet's performance suggests an inclination toward overfitting, underscoring the importance of balancing model complexity with predictive accuracy in regression tasks. Overall, the varying performance metrics underscore the necessity for model-specific optimizations and highlight the trade-offs between model complexity and prediction accuracy.

\section{Results and Analysis}
Based on the comparative analysis of the RSME validation scores of different machine learning models presented in section III, we chose XGBoost and LightGBM models for extracting the material parameters of the inkjet printed CPWs.

\looseness=-1 Once the training was completed, we proceeded for extracting the material parameters from measured propagation constant. To this end, the attenuation, phase constant and frequencies were provided as input to the proposed AutoML model. Fig. 7 depicts the frequency dependent material parameters predicted from our proposed AutoML architecture. It can be seen that predicted $\epsilon_{DS}$ vary between 1.7 and 2.2; whereas $\epsilon_{FS}$ vary between 2.6 and 3.4. The predicted loss tangent of the flexible substrate is within 0.008 and 0.013. Finally, the conductivity of cu electroplated silver nanoparticle ink vary between $2.8\times 10^7$ S/m and $3.2\times 10^7$ S/m. 
The extreme outliers for each prediction can be attributed to the worst case modeling error for the AutoML architecture and can be ignored. Nevertheless, we used the mean value from Fig. 7 to estimate the material parameters. 

\begin{figure}[t]
	\centering
	\includegraphics[width=0.95\columnwidth]{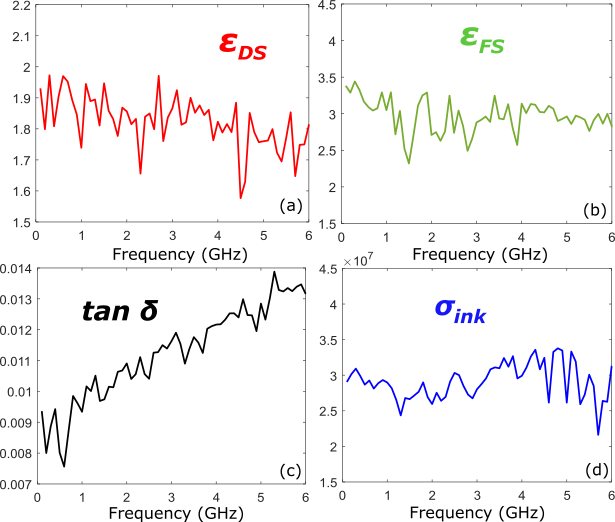}
	\caption{\small{Extracted material parameters from the proposed automated machine learning model: (a) $\epsilon_{DS}$ (b) $\epsilon_{FS}$ (c) $tan \delta$ (d) $\sigma_{ink}$. }}
	\label{f:Test_figure2}
\end{figure}

\begin{table}[t]
	\renewcommand{\arraystretch}{1.3}
	\centering
	\caption{Comparison of material parameters extracted from conventional method and predicted from proposed machine learning model} 
	\begin{tabu} to \columnwidth { |X[c]| X[c]| X[c]|}
		\tabucline[0.6pt]{-}
		{\textbf{Parameters}} &  \textbf{Measured} &  \textbf{Predicted}\\
		\tabucline[0.6pt]{-}
		$\pmb{\epsilon_{FS}}$ & 3.2 & 3.05\\
		\hline
		$\pmb{\epsilon_{DS}}$ & 1.81  & 1.85\\
		\hline
		$\pmb{tan \delta}$ & 0.01   & 0.0113\\
        \hline
        $\pmb{\sigma_{ink} (10^7}$\textbf{S/m)} & 2.973 & 3.107\\
		\tabucline[0.6pt]{-}
	\end{tabu}
\end{table}

\begin{figure}[t]
	\centering
	\includegraphics[width=.9\columnwidth]{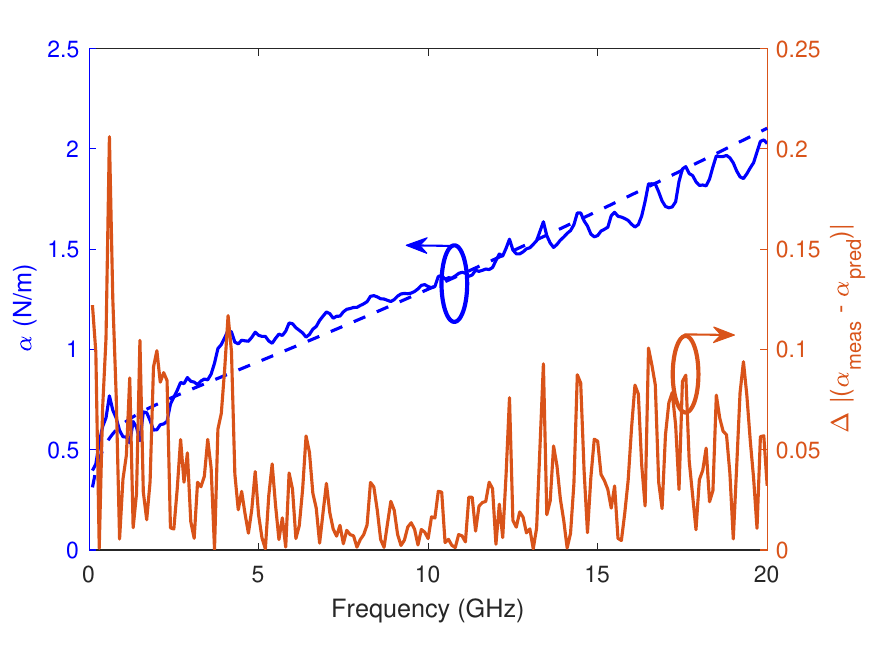}
	\caption{\small{Comparison of attenuation constant obtained from multiline TRL and EM simulation with material parameters predicted from our proposed machine learning model. The difference between measured and predicted attenuation constant is plotted on the right.}}
	\label{f:Test_figure2}
\end{figure}

\looseness=-1 In order to verify the retrieved material parameters are correct, we measured measured $\sigma_{ink}$, $\epsilon_{FS}$, $\epsilon_{DS}$, $tan \delta$ using alternative approaches. 
A split postresonator method was adopted below 5 GHz and a waveguide technique was adopted from 18 to 27 GHz to measure $\epsilon_{DS}$. The average value from both methods was 1.81. The measured values for dielectric constant and loss tangent of the flexible substrate, $\epsilon_{FS}$ and $tan \delta$ were adopted from literature \cite{c1} as 3.2 and 0.01, respectively. Finally, the dc conductivity of the nanoparticle ink was measured using 
$\sigma_{ink}^{dc}= \dfrac{l}{R\times A}$,
where we measured the dc resistance of printed CPW lines
using an LCR meter connected to the VNA as shown in Fig. 1. The measured dc conductivity was found to be $2.973\times 10^7$ S/m. Table 1 compares the measured material parameters with the mean of frequency dependent material parameters predicted from our proposed AutoML architecture. As can be seen both values agree quite well with each other demonstrating the effectiveness of our approach.

\looseness=-1 As an additional verification, we performed an finite-element simulation using the retrieved material parameters from table 1. Fig. 8 shows  comparison of attenuation constant from EM simulation with extracted parameters (in dashed blue line) and measurement (in solid blue line). We also plot the absolute difference between simulated and measured attenuation constant (in solid orange line) in Fig. 5. As can be observed, the worst case difference for the attenuation constant is around 0.2 N/m, demonstrating the effectiveness of our proposed methodology. As can be seen, we are able
to extract the values of material parameters using only a single set of measurements. This suggests that our proposed approach may be of use in process monitoring for additive manufacturing. 

\section*{Conclusion}

\looseness=-1 An automated machine learning based architecture has been proposed for smart characterization of inkjet printed components. Our proposed architecture evaluates numerous state-of-the-art machine learning models and automatically selects the best algorithm for extraction of each parameter. Through extensive training and validation, we showed that XGBosst and LightGBM methods offer better accuracy for the modeling and regression task under study. The proposed characterization methodology was able to accurately predict the  ink conductivity and the dielectric properties of the flexible substrate and spacer used for measurement. The employment of machine learning based models enable full automation of the extraction procedure. Furthermore, the proposed AutoML architecture eliminates the human expertise required to select appropriate algorithms. The extracted material parameters were found to be in good agreement with the measured values.

\end{document}